\documentclass[journal,twoside,web]{ieeecolor}
\usepackage{etoolbox}
\makeatletter
\@ifundefined{color@begingroup}%
{\let\color@begingroup\relax
	\let\color@endgroup\relax}{}%
\def\fix@ieeecolor@hbox#1{%
	\hbox{\color@begingroup#1\color@endgroup}}
\patchcmd\@makecaption{\hbox}{\fix@ieeecolor@hbox}{}{\FAILED}
\patchcmd\@makecaption{\hbox}{\fix@ieeecolor@hbox}{}{\FAILED}
\usepackage{tmi}
\usepackage{cite}
\usepackage{amsmath,amssymb,amsfonts}
\usepackage{algorithmic}
\usepackage{graphicx}
\graphicspath{{fig/},{pics/}}
\usepackage{subfigure}
\usepackage{textcomp}
\usepackage{multirow}
\usepackage{booktabs}
\usepackage{bm}
\usepackage{bbding}
\usepackage{amssymb}
\usepackage{amsmath}

\usepackage{color}

\makeatletter
\let\NAT@parse\undefined
\makeatother
\usepackage{hyperref}
\hypersetup{hidelinks}

\usepackage{algorithm}
\usepackage{algorithmic}

\makeatletter
\newcommand{\rmnum}[1]{\romannumeral #1}
\newcommand{\Rmnum}[1]{\expandafter\@slowromancap\romannumeral #1@}
\makeatother

\usepackage[T1]{fontenc} 

\begin{document}
	\title{HST-MRF: Heterogeneous Swin Transformer with Multi-Receptive Field for Medical Image Segmentation}
	\author{Xiaofei Huang, Hongfang Gong, Jin Zhang
		\thanks{This work was supported in part by the National Natural Science Foundation of China under Grant 61972055, and in part by the Natural Science Foundation of Hunan Province under Grant 2021JJ30734. (Corresponding author: Hongfang Gong.)}
		\thanks{Xiaofei Huang, Hongfang Gong are with the school of Mathematics and Statistics, Changsha University of Science and Technology, Changsha 410114, China (e-mail:xiaofeihuang2021@163.com; ghongfang@126.com).}
		\thanks{Jin Zhang is with the school of Computer and Communication Engineering, Changsha University of Science and Technology, Changsha 410114, China (e-mail:mail\_zhangjin@163.com).}}
	
	\maketitle
	
	\begin{abstract}
		The Transformer has been successfully used in medical image segmentation due to its excellent long-range modeling capabilities. However, patch segmentation is necessary when building a Transformer class model. This process may disrupt the tissue structure in medical images, resulting in the loss of relevant information. In this study, we proposed a Heterogeneous Swin Transformer with Multi-Receptive Field (HST-MRF) model based on U-shaped networks for medical image segmentation. The main purpose is to solve the problem of loss of structural information caused by patch segmentation using transformer by fusing patch information under different receptive fields. The heterogeneous Swin Transformer (HST) is the core module, which achieves the interaction of multi-receptive field patch information through heterogeneous attention and passes it to the next stage for progressive learning. We also designed a two-stage fusion module, multimodal bilinear pooling (MBP), to assist HST in further fusing multi-receptive field information and combining low-level and high-level semantic information for accurate localization of lesion regions. In addition, we developed adaptive patch embedding (APE) and soft channel attention (SCA) modules to retain more valuable information when acquiring patch embedding and filtering channel features, respectively, thereby improving model segmentation quality. We evaluated HST-MRF on multiple datasets for polyp and skin lesion segmentation tasks. Experimental results show that our proposed method outperforms state-of-the-art models and can achieve superior performance. Furthermore, we verified the effectiveness of each module and the benefits of multi-receptive field segmentation in reducing the loss of structural information through ablation experiments.
	\end{abstract}
	
	
	\section{Introduction}
	\IEEEPARstart{M}{edical} image segmentation is a critical and complex step in medical image analysis\cite{medical-image-segmentaion}. The task aims to segment lesion and nonlesion regions in the image. Effective medical image segmentation models can assist medical professionals in improving diagnostic efficiency and are an important component of intelligent healthcare systems and computer-aided diagnosis\cite{mis}. Deep learning has been widely applied to medical image segmentation tasks, including but not limited to vessel detection\cite{vessel1}\cite{vessel2}, brain segmentation\cite{brain1}\cite{brain2}, optic disc segmentation\cite{disc1}\cite{disc2}, and lung segmentation\cite{lung1}\cite{lung2}, due to its remarkable performance.
	
	Convolutional neural networks (CNNs) are popular among researchers in computer vision because of their simple operations and efficient feature extraction capability. Many related excellent CNN models, such as Fully Convolutional Network (FCN)\cite{fcn} and U-Net\cite{unet}, have been proposed for image segmentation tasks. These model rely on the encoder-decoder structure. The encoder is used to extract the high-level semantic information of the image, and the decoder is used to recover the resolution of the feature map step by step. With the help of skip connection, the information of the corresponding stages is combined to achieve precise targeting. Many current derivation methods based on both, such as UNet++\cite{unet++} and mU-Net\cite{munet}, have achieved great success in medical image segmentation. However, the locality and translation invariance of CNNs fail to model long-range dependency of visual information in images although they can give advantages to local modeling and computation. Therefore, the performance of these methods is obviously insufficient when dealing with images with large differences in target structures\cite{transunet}.
	
	Transformer\cite{attention} have brought breakthroughs in the field of natural language processing due to their excellent ability to model global context. To this end, numerous studies have explored how to apply them to the field of computer vision in recent years. DEtection TRansformer (DETR)\cite{detr} builds an end-to-end target detection model based on Transformer, views each pixel in the feature map as a token, and explores their global relationship. Vision Transformer (ViT)\cite{vit} divides the image into a patch sequence and uses it as the input of the Transformer, so that it can be directly applied to the entire image. ViT has achieved good performance in image classification tasks. TransUNet\cite{transunet} uses Transformer encoding on the basis of U-Net and obtains more powerful context features and better segmentation accuracy. Zhang \textit{et al.} proposed TransFuse\cite{transfuse}, which uses Transformer and CNNs to extract the global context information and local detail information of the image, respectively, and effectively combines the coding information of the two branches through the BiFusion module. These methods successfully apply Transformer to vision tasks and solve the problem that CNNs cannot model long-range dependency in image processing.
	
	In the process of using the visual Transformer, the image needs to be divided into patches, which can damage the structure information of the tissue in the medical image and seriously affect the performance of model. With the transformation of medical image segmentation tasks, the size, shape, and location of lesions vary greatly, and the fixed-size receptive field patch segmentation method cannot adapt to this change. Lin \textit{et al.} proposed DS-TransUNet\cite{dstransunet}, a Swin Transformer\cite{ST} based model for medical image segmentation tasks that incorporates parallel dual-scale encoding and a Transformer Interactive Fusion (TIF) module for complementary encoding information across different scale patches. Although this design reduces structural information loss, the TIF module is not utilized during the encoding process and cannot propagate the complementary information across stages, limiting its effectiveness. When handling image classification tasks, CrossViT\cite{crossvit} utilizes cross-attention to fuse image blocks of different scales to obtain a stronger representation of image features. HRViT\cite{hrvit} enhances the learning ability of ViT through the high-resolution multibranch structure and learns rich multiscale representations to solve the semantic segmentation problem. These approaches can alleviate the drawbacks of patch segmentation by fusing multiscale information. However, blindly computing the correlation between all different scales of lesion regions can be computationally expensive and may introduce correlation confusion problems.
	
	Ke \textit{et al.} proposed a method to calculate word-relatedness and position-relatedness using different projection matrices to introduce better positional bias into the Transformer\cite{tupe}. Inspired by this, we propose the Heterogeneous Swin Transformer with Multi-Receptive Field (HST-MRF), a U-shaped network structure that uses heterogeneous attention computation on multi-receptive field features of the image to compensate for the limitations of patch segmentation. We designed the heterogeneous Swin Transformer (HST) module to guide the interactive learning of multi-receptive field features and model the long-range dependencies among elements, which are then passed on to the next stage for progressive learning to achieve structural information complementarity. Specifically, at each stage, we calculate the elemental correlations of the feature maps under different receptive fields using different parameterizations and aggregate them into an overall attention score function. Based on this function, we model the long-range relationships among the feature maps. The two-stage fusion module, multimodal bilinear pooling (MBP), is designed to assist HST in further fusing multi-receptive field information and effectively combining the low-level and high-level features of the corresponding stages for precise lesion localization. We also designed the adaptive patch embedding (APE) and soft channel attention (SCA) modules to further improve segmentation quality through adaptive computation.
	
	Our contributions can be summarized as:
	\begin{itemize}
		\item HST-MRF model is proposed, which addresses the issue of structural information loss caused by patch segmentation when using Transformer.
		\item We design the APE and HST modules to model the long-range dependencies of visual information and effectively fuse patch information under different receptive fields. The MBP and SCA modules are introduced to fuse and filter low-level and high-level feature information.
		\item We evaluate HST-MRF on datasets related to polyp and skin lesion segmentation and showed that HST-MRF outperforms related methods. We also validate the effectiveness of each module and multi-receptive field patch segmentation by ablation experiments.
	\end{itemize}
	
	The rest of the paper is structured as follows. Section \Rmnum{2} reviews the related work. Section \Rmnum{3} describes HST-MRF in detail. Section \Rmnum{4} performs a series of extensive experiments and ablation study on our proposed model. Section \Rmnum{5} discusses our work and provides the conclusions.

	\section{Related Work}
	
	\subsection{CNNs for Medical Image Segmentation}
	The encoder-decoder structure has been widely used in various image segmentation tasks due to its success in pixel-level image segmentation. The U-Net\cite{unet}, which has an encoder-decoder structure, was proposed for 2D medical image segmentation and demonstrated excellent performance, opening up a new research direction for medical image segmentation. However, the skip connection in U-Net may fuse different semantic information of the encoder and decoder, causing feature confusion. UNet++\cite{unet++} was proposed to solve this problem by designing a dense skip connection. Ning \textit{et al.} proposed SMU-Net\cite{smunet} to use the background of the image to assist the foreground in segmentation when dealing with breast ultrasound image segmentation. Gu \textit{et al.} applied the idea of Inception\cite{inception} to U-Net and proposed CE-Net\cite{cenet} to solve the problem of spatial information loss due to cross-step convolution and pooling in U-Net and its variants. Feng \textit{et al.} designed CPFNet\cite{cpfnet} to address the problem of insufficient extraction capability of single stage contexts of U-shaped networks, which effectively fuses global and local context information through GPG and SAPF modules.
	
	\begin{figure*}[h]
		\centering
		\includegraphics[width=0.85\textwidth]{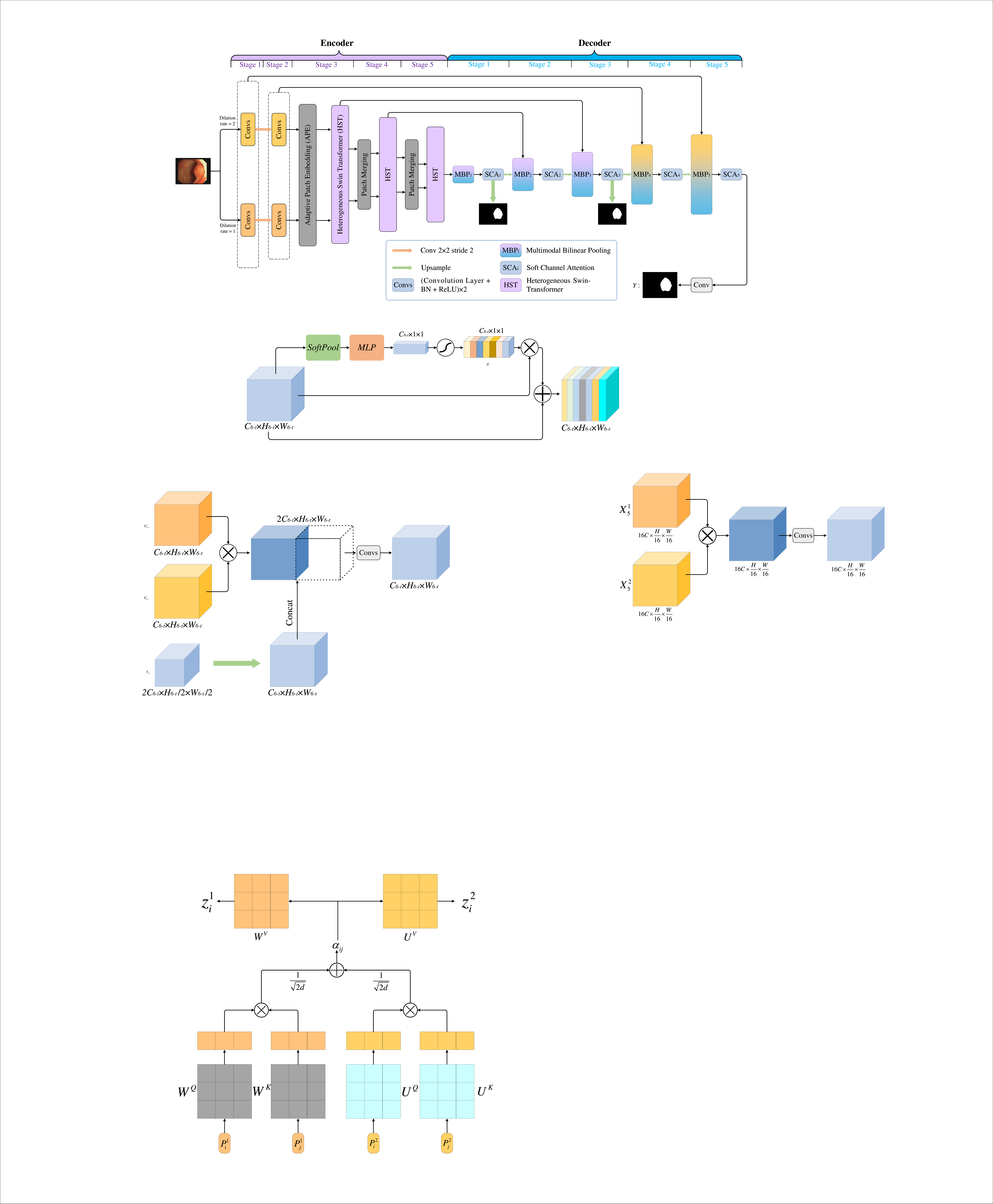}
		\caption{Our proposed heterogeneous Swin Transformer with multiple receptive fields (HST-MRF). We utilized APE and HST modules in the encoding process, and five MBP modules (${\rm MBP}_1$ to ${\rm MBP}_5$), five SCA modules (${\rm SCA}_1$ to ${\rm SCA}_5$) in the decoding process.}
		\label{fig1}
	\end{figure*}

	In addition to CNNs-based U-Net variant, recent studies have introduced the attention mechanism to help CNNs capture important information. Attention-Unet\cite{attentionunet} adds an attention-gating mechanism to U-Net, allowing the model to focus on salient regions in the image and suppress irrelevant regions. Tomar \textit{et al.} proposed FANet\cite{fanet}, which strictly attends to the subsequent training process by utilizing the prediction maps obtained in each training epoch during medical image segmentation. CA-Net\cite{canet} solves the problems of object spatial position, channel redundancy, and object scale correlation of segmented targets in medical images by spatial attention, channel attention, and scale attention, respectively. These schemes have shown excellent performance in medical image segmentation tasks. However, they are all based on CNNs, which lack the ability to model the long-range correlation of visual information in images.

	\subsection{Transformer for Image Segmentation}
	The structure of Transformer can model the long-range relationships between sequence elements, leading many researchers to explore its application in solving the above problem. Carion \textit{et al.} first applied Transformer to vision tasks and developed DETR\cite{detr} for object detection. Subsequently, Dosovitskiy \textit{et al.} designed ViT\cite{vit}, keeping the original structure of Transformer unchanged and using the patches into which the given image is segmented as input. Although they achieve good results in visual long-range relationship modeling, they require the image to be divided into smaller patches when dealing with intensive prediction tasks, such as semantic segmentation, which increases the computational cost. To solve this problem, Liu \textit{et al.} designed window attention and shifted window mechanisms and developed Swin Transformer\cite{ST}, which improves the efficiency of cross-window visual information modeling.
	
	For image segmentation tasks, SETR\cite{setr} extracts image features using Transformer as an encoder rather than the commonly used stacked convolution. A simple decoder is then used to obtain semantic segmentation results, converting the vision encoding-decoding task to sequence-to-sequence. Segmenter\cite{segmenter} proposed by Strudel \textit{et al.} is an encoder-decoder structure based entirely on Transformer, using ViT to encode images, and proposes mask transformer to decode the output of the encoder and class embedding. SegFormer\cite{segformer} is a simple and efficient semantic segmentation framework that can obtain feature outputs at different scales through a hierarchical Transformer encoder. A decoder composed of MLP aggregates the information of each layer to show a more powerful expressiveness. MCTrans\cite{mctrans} learns dependencies between pixels across scales in medical images by integrating semantic structure mining and rich feature learning into a single framework.	
	
	Although these schemes achieve decent performance in image segmentation tasks, the necessity of patch segmentation destroys the structural information, which may be fatal to medical image segmentation tasks. To address this issue, we propose an interactive encoding mechanism based on Swin Transformer that can complement different receptive field patch information and reduce the loss of structural information.	
	
	\section{Methods}
	The HST-MRF framework, consisting of four key components: APE, HST, MBP, and SCA, is illustrated in \textcolor{mblue}{Fig. \ref{fig1}}. HST utilizes heterogeneous attention to enable interaction between feature maps of different receptive fields, which are then propagated to the next stage for progressive computation. MBP further assists HST in fusing multi-receptive field information and combining feature information corresponding to the encoder-decoder. APE and SCA preserve more relevant information through adaptive computation in the patch embedding and channel feature selection processes, respectively, resulting in more accurate segmentation predictions.
	
	\subsection{Encoder}	
	Unlike the traditional U-Net encoder, we design an interactive encoding mechanism with multi-receptive fields. The mechanism comprises five stages, and the output features from each stage result from the interaction of different receptive fields. Let $I\in{\mathbb R}^{3\times H\times W}$ denote the input medical image, where $H\times W$ represents the spatial resolution of the input image. The encoder includes APE and HST modules.
	
	\subsubsection{APE}
	To ensure that same-scale patches have different receptive fields for information complementarity, we use dilated convolutions with a kernel size of 3 and dilation rates of 1 and 2 to perform initial encoding on $I$ in the first and second stages, respectively. We also used $2\times2$ convolution with a step size of 2 to reduce the resolution of the feature maps. The outputs of different receptive fields in the two stages are $X^1_1,X^2_1\in{\mathbb R}^{C\times H\times W}$ and $X^1_2, X^2_2\in{\mathbb R}^{2C\times \frac{H}{2}\times \frac{W}{2}}$, where $X^r_m$ represents the output feature map of the expansion rate $r$, the $m$-th encoding stage.
	
	Feature maps are often projected into several 1D vectors using a linear projection to enable Transformer to model visual information globally. Although this operation is simple, it is not flexible enough to handle different regions of the image. Motivated by SoftPool\cite{softpool}, we designed the APE module with two operation stages. In the first stage, the segmentation patch is transformed into a patch vector using SoftPool, where each component represents an independent channel feature. In the second stage, we performed linear operations on the patch vectors obtained from the previous step to complete feature combination, and obtain the final patch embedding.
	
	In the first stage, we first divide the feature maps $X^1_2$ and $X^2_2$ into $N=\frac{H}{2S}\times \frac{W}{2S}$ nonoverlapping patches, where $S=2$ is the size of the patch. Then, we use SoftPool to derive a vector representation of each patch. The calculation can be expressed as follows:
	
	\begin{equation}
		\begin{aligned}
			p^r_{c,i} &= \sum_{s}^{S^2}\frac{\exp(a^r_{c,i,s})a^r_{c,i,s}}{\sum_{s'=1}^{S^2}\exp(a^r_{c,i,s'})},\\
			p^r_i&=[p^r_{1,i},p^r_{2,i},\cdots,p^r_{2C,i}],		
		\end{aligned}
	\end{equation}
	where $p^r_{c,i}\in{\mathbb R}^{2C}$ can be regarded as the pixel representation of the $i\in\{1,2,\cdots,N\}$-th patch in the $c\in\{1,2,\cdots,2C\}$-th channel with dilation rate $r$, and $a^r_{c,i,s}$ denotes the dilation rate $r$, the pixel value of the $i$-th patch at position $s$ in the $c$-th channel. $p^r_i$ is the $i$-th patch vector representation under dilation rate $r$. 
	
	In the second stage, the patch vectors are fed into a fully connected layer with an output dimension of $4C$ (${\rm FC}_{(4C)}$) to combine features and obtain a more expressive patch representation for the dilation rate $r$, that is, $P^r={\rm FC}_{(4C)}(p^r)$, where $p^r=[p^r_1,p^r_2,\cdots,p^r_N]$, $P^r=[P^r_1,P^r_2,\cdots,P^r_N]$, and $P^r_i$ denote the embedding representation of the $i$-th patch with dilation rate $r$. From this, we can obtain the final patch embedding representation under the two receptive fields as $P^1\in{\mathbb R}^{4C\times N}$ and $P^2\in{\mathbb R}^{4C\times N}$. APE can flexibly handle features from different regions of the image by utilizing SoftPool, preserving channel features better and improving segmentation quality.
	
	\subsubsection{HST}
	\begin{figure}[h]
		\centering
		\centerline{\includegraphics[width=0.75\columnwidth]{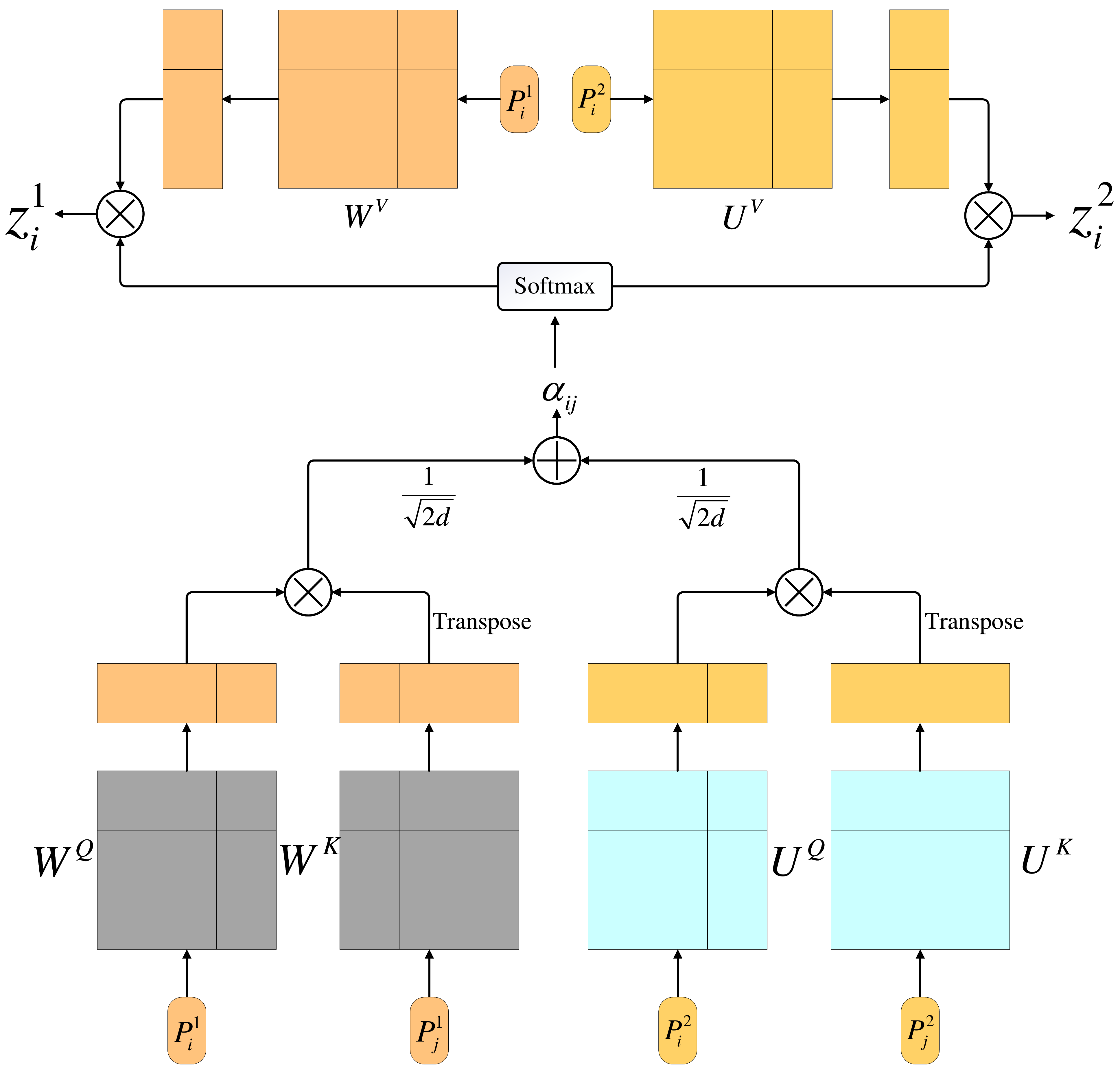}}
		\caption{Description of heterogeneous attention computation in HST.}
		\label{fig2}
	\end{figure}
	Transformer can effectively model long-range dependencies of visual information. However, a problem of losing organizational structure information due to patch segmentation exists. Fusing patches with different receptive fields can alleviate this issue. Therefore, exploring effective methods to integrate patch feature information from different receptive fields is important. To address this, we propose the HST module based on Swin Transformer that interacts information from multiple receptive fields while modeling long-range relationships to reduce the loss of image structure information.

	\textcolor{mblue}{Fig. \ref{fig2}} shows the computation of heterogeneous attention in HST. We calculate the correlation between patches under each receptive field using different projection matrices and sum them as the overall attention distribution,
	\begin{equation}
		\resizebox{.9\hsize}{!}{$ \alpha_{ij}=\frac{1}{\sqrt{2d}} \left((P^1_i{\bf W}^Q)(P^1_j{\bf W}^K)^T+(P^2_i{\bf U}^Q)(P^2_j{\bf U}^K)^T\right)$},
	\end{equation}
	where $d$ represents the dimension of the model's hidden layer. $\alpha_{ij}$ denotes the overall correlation between patches $i$ and $j$ obtained by combining information from all corresponding positions of different receptive fields. Thus, it can be obtained that,
	\begin{equation}
		\begin{aligned}
			z^1_i=&\sum_{j=1}^{n_1}\frac{\exp(\alpha_{ij})}{\sum_{j'=1}^{n_1}\exp(\alpha_{ij'})}(P^1_i {\bf W} ^V),\\ z^2_i=&\sum_{j=1}^{n_2}\frac{\exp(\alpha_{ij})}{\sum_{j'=1}^{n_2}\exp(\alpha_{ij'})}(P^2_i{\bf U}^V),
		\end{aligned}
		\label{eq3}
	\end{equation}
	where $z^1_i$ and $z^2_i$ denote the dot product outputs of patch $i$ under two dilation rates, and $n_1$ and $n_2$ represent the number of patches in the two receptive fields, where $n_1=n_2$. The computation of heterogeneous attention is formulated as:
	\begin{equation}
		\resizebox{0.9\hsize}{!}{$
			\begin{aligned}
				{\rm HA}(Q,K,V)=&{\rm Softmax}\left(\frac{Q_1K_1^T+Q_2K_2^T}{\sqrt{2d}}\right)V\\
				=&[z^1_1,\cdots,z^1_{n_1};z^2_1,\cdots,z^2_{n_2}],
			\end{aligned}$}
	\end{equation}
	where $Q=[Q_1;Q_2]$, $K=[K_1;K_2]$, and $V=[V_1; V_2]$ are the concatenation of the Query, Key and Value matrices under two receptive fields, respectively.
	
	We combine the above heterogeneous attention with Swin Transformer to obtain the W-HMSA and SW-HMSA operations of the HST module, and the calculation of W-HMSA can be expressed as	\begin{equation}
		\begin{aligned}
			\resizebox{0.9\hsize}{!}{${\rm W-HMSA}(Q,K,V)=[{\rm head}_1;{\rm head}_2;\cdots;{\rm head}_{n_h}]{\bf W}^o,$}\\
			{\rm head}_j={\rm HA}(Q{\bf W}^Q_j,K{\bf W}^K_j,V{\bf W}^V_j),
		\end{aligned}
	\end{equation}
	where $Q\in{\mathbb R}^{d\times 2hw}$, $K\in{\mathbb R}^{d\times 2hw}$, $V\in{\mathbb R}^{d\times 2hw}$. Here $h$, $w$ denote the size of the window in Swin Transformer. ${\bf W}^Q_j\in{\mathbb R}^{2hw\times 2d_q}$, ${\bf W}^K_j\in{\mathbb R}^{2hw\times 2d_k}$, ${\bf W}^V_j\in{\mathbb R}^{2hw\times 2d_v}$, and ${\bf W}^o\in{\mathbb R}^{2n_hd_v\times 2d}$ are learnable block diagonal projection matrices, where $n_h$ is the number of heads, and $d_q=d_k=d_v$ are the dimensions of query, key and values, respectively. Adding a shift window to W-HMSA, we can obtain SW-HMSA. 
	
	With HST, we combine the information interaction of multi-receptive fields with Swin Transformer and propagate it to the next stage for progressive learning, achieving complementary structural information. We also progressively increase the receptive field of the feature map by a patch merging layer (reducing resolution by 2$\times$ and increasing channels by 2$\times$) to obtain high-level semantic information of the image $I$. As shown in \textcolor{mblue}{Fig. \ref{fig1}}, we use patch merging and HST in the fourth and fifth stages.

	The output feature maps of the encoder in five stages under two receptive fields can be represented as: $X^1_1,X^2_1\in{\mathbb R}^{C\times H\times W}$, $X^1_2,X^2_2\in{\mathbb R}^{2C\times\frac{H}{2}\times \frac{W}{2}}$, $X^1_3,X^2_3\in{\mathbb R}^{4C\times\frac{H}{4}\times \frac{W}{4}}$, $X^1_4,X^2_4\in{\mathbb R}^{8C\times\frac{H}{8}\times \frac{W}{8}}$, $X^1_5,X^2_5\in{\mathbb R}^{16C\times\frac{H}{16}\times \frac{W}{16}}$.
	
	\subsection{Decoder}
	The main role of the decoder is to gradually recover the resolution of the feature map based on the high-level semantic information of the image and obtain pixel-level segmentation results. Each stage of decoder contains the MBP and SCA modules, see \textcolor{mblue}{Fig. \ref{fig1}}.
	
	\subsubsection{MBP}
	\begin{figure}[h]
		\centering
		\centerline{\includegraphics[width=0.8\columnwidth]{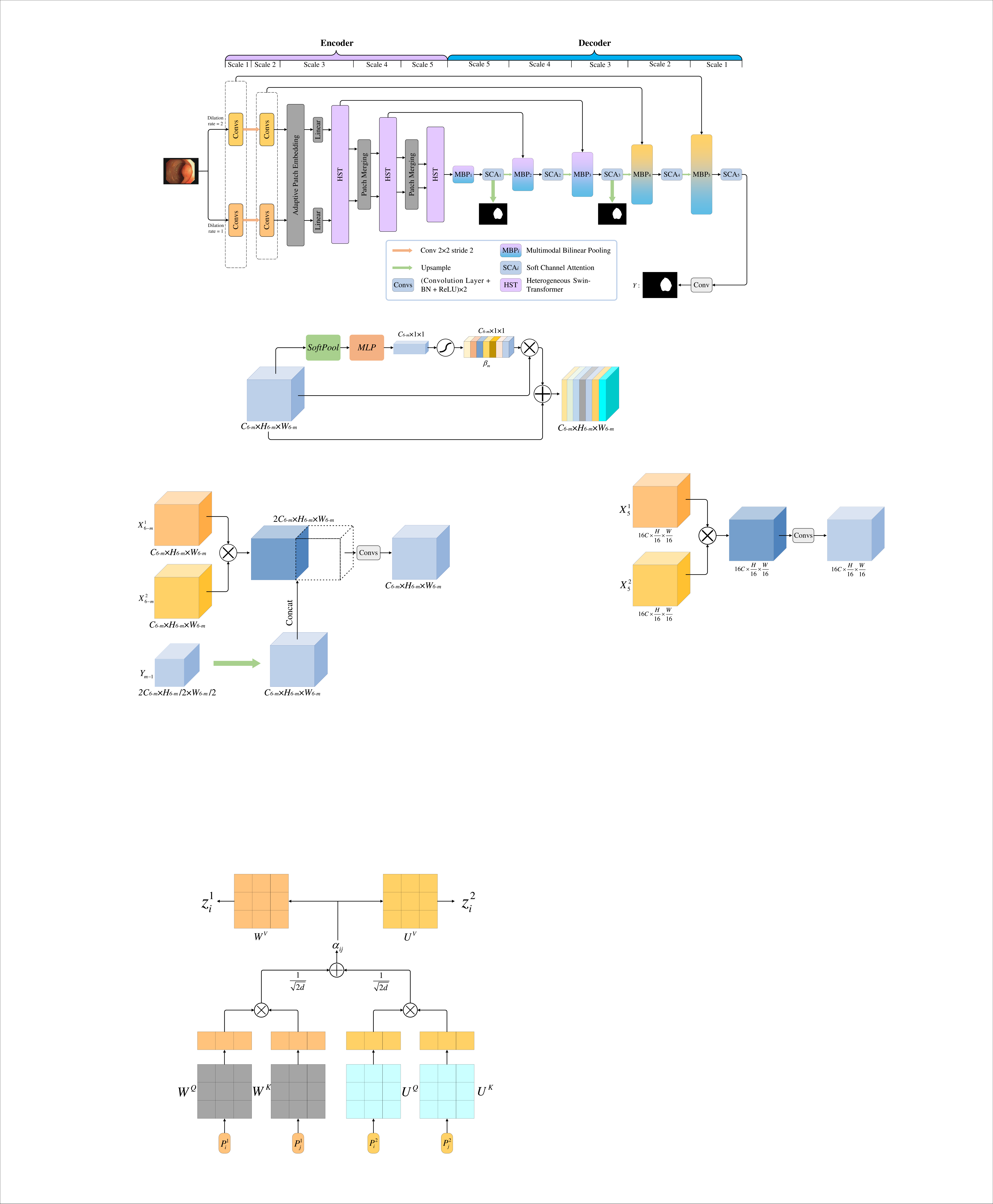}}
		\caption{The operation process of ${\rm MBP}_1$. First, we perform Hadamard product between $X^1_5$ and $X^2_5$, and then obtain the output $D_1$ of ${\rm MBP}_1$ through Convs.}
		\label{fig3}
	\end{figure}
	Skip connection is a common method for accurately localizing lesion regions. However, we obtain two sets of encoded information in each stage, and directly fusing them using skip connection may result in feature confusion issues, similar to that reported by Wang \textit{et al.}\cite{UCTransNet}. Inspired by the fact that bilinear pooling can be effective for the fusion of multimodal information, we designed a two-stage fusion module, MBP. In MBP, we first fuse the encoded low-level semantic information and then further merge the high-level semantic information. Specifically, we use five MBP modules (${\rm MBP}_t$, $t\in\{1,2,3,4,5\}$) for information fusion at different resolution levels.

	When $t=1$, the inputs of ${\rm MBP}_1$ consists of high-level semantic features $X^1_5$ and $X^2_5$. The computation of ${\rm MBP}_1$ can be expressed as follows:
	\begin{equation}
		D_1={\rm MBP}_1(X^1_5,X^2_5)={\rm Convs}(X^1_5\circ X^2_5),
	\end{equation}
	where $\circ$ denotes Hadamard product, $D_1$ denotes the output of ${\rm MBP}_1$, and Convs is composed of two consecutive layers of Convolution, BatchNorm (BN) and ReLU, which mainly function to enrich the features after fusion. The calculation process of ${\rm MBP}_1$ is shown in \textcolor{mblue}{Fig. \ref{fig3}}.
	
	\begin{figure}[h]
		\centering
		\centerline{\includegraphics[width=0.8\columnwidth]{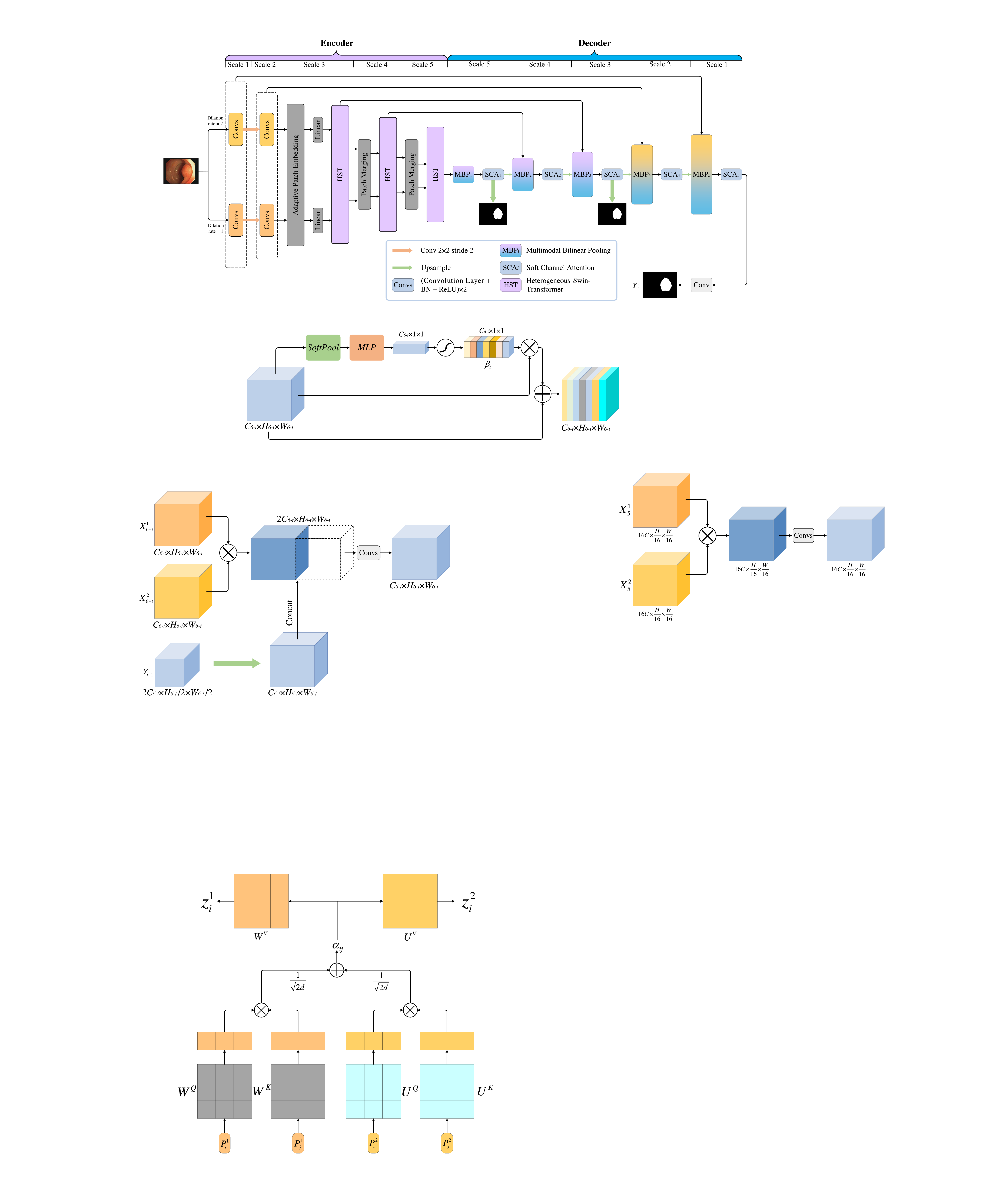}}
		\caption{The operation process of ${\rm MBP}_t$ ($t>1$). First, we obtain the overall low-level semantic information $X^1_{6-t}\circ X^2_{6-t}$, and the upsampled output $Y_{t-1}$, ${\rm Up}(Y_{t-1})$. Then, we concatenate them and apply Convs to get the output $D_t$ of ${\rm MBP}_t$.}
		\label{fig4}
	\end{figure}
	
	When $t>1$, the input of ${\rm MBP}_t$ consists of three parts, namely, the encoded information from two receptive fields and the previous decoder stage output. The computation of ${\rm MBP}_t$ can be represented as follows:
	\begin{equation}
		\resizebox{0.9\hsize}{!}{$D_t={\rm MBP}_t(X^1_{6-t},X^2_{6-t},Y_{t-1})={\rm Convs}([X^1_{6-t}\circ X^2_{6-t};{\rm Up}(Y_{t-1})])$},
	\end{equation}
	where $Y_{t-1}$ denotes the output of the $(t-1)$-th stage of the decoder, which is upsampled to ensure the same resolution as the output feature map of the corresponding encoder and concatenated with overall low-level semantic information $X^1_{6-t}\circ X^2_{6-t}$. Convs is then used to enrich the features and obtain the output $D_t$ of ${\rm MBP}_{t}$. The entire process is shown in \textcolor{mblue}{Fig. \ref{fig4}}, where $C_{6-t}$, $H_{6-t}$, and $W_{6-t}$ represent the number of channels, height, and width of the output feature maps from the $(6-t)$-th encoding stage, respectively.

	Through the two-stage fusion of MBP, we facilitate HST to further merge patch information from different receptive fields and achieve secondary complementary structure information, and effectively combine semantic information from encoding and decoding to reduce the possibility of feature confusion and achieve more accurate segmentation.	
	
	\subsubsection{SCA}
	\begin{figure}[h]
		\centering
		\centerline{\includegraphics[width=0.8\columnwidth]{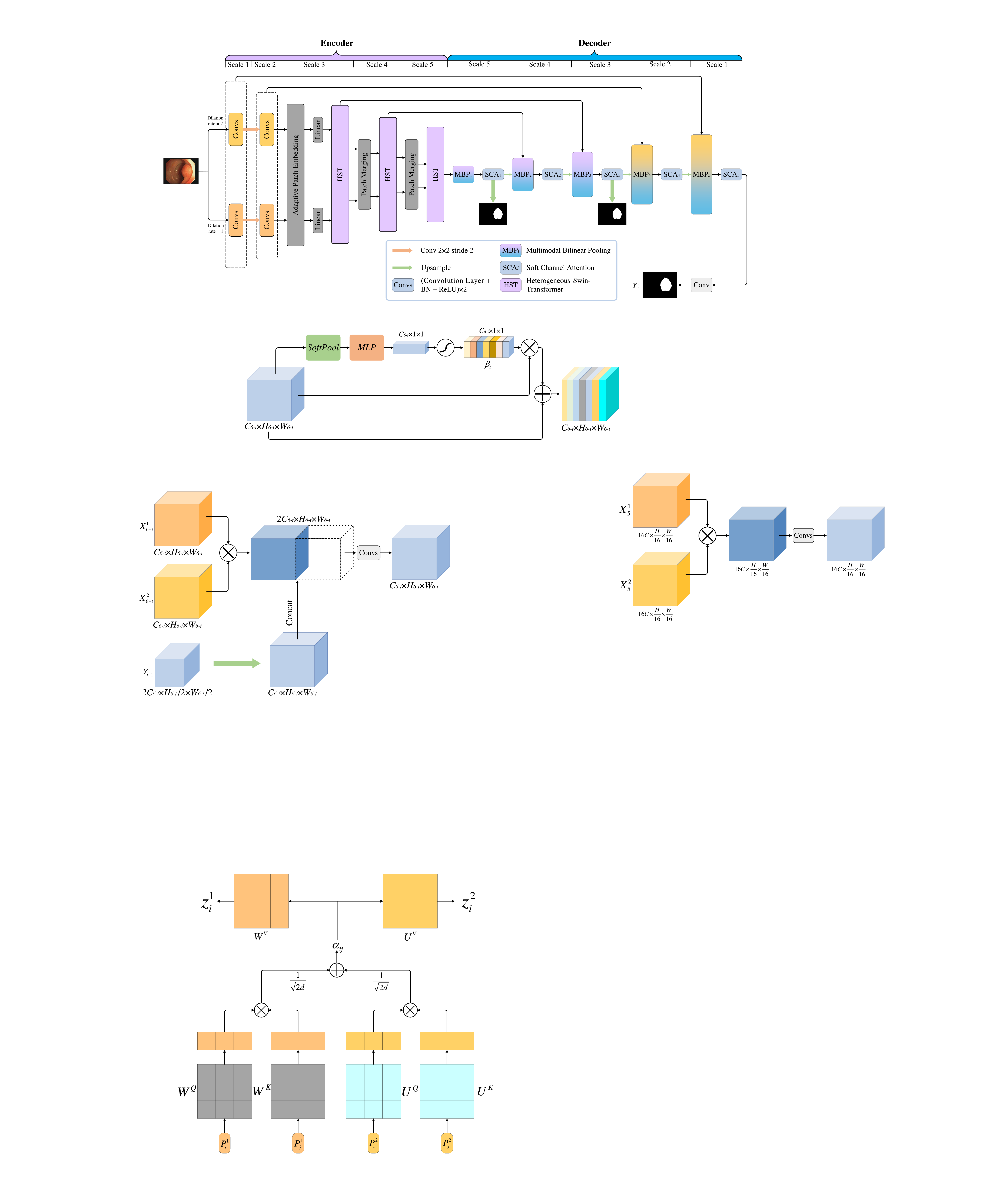}}
		\caption{The operation process of ${\rm SCA}_t$. Based on the feature map $D_t$, the attention weight $\beta_t$ is obtained for each channel, and then the output $Y_t$ of the ${\rm SCA}_t$ is obtained through residual connections.}
		\label{fig5}
	\end{figure}
	The MBP module effectively combines information from the encoder and decoder, but channel redundancy may occur inevitably during computation, resulting in noise. To address this, we designed the ${\rm SCA}$ module based on SoftPool (see \textcolor{mblue}{Fig. \ref{fig5}}), denoted as ${\rm SCA}_t(t\in\{1,2,3,4,5\})$, with the input being the output of the corresponding stage MBP. SoftPool is used to obtain the global channel information of $D_t$, that is, ${\rm SoftPool}(D_t)=[d_{t,1}, d_{t,2},\cdots, d_{t,C_{6-t}}]$, where $d_{t,c}$ is calculated as follows:
	\begin{equation}
		d_{t,c}=\sum_{s=1}^{H_{6-t}W_{6-t}}\frac{\exp{(b_{t,c,s})}b_{t,c,s}}{\sum_{s'=1}^{H_{6-t}W_{6-t}}\exp{(b_{t,c,s'})}},
	\end{equation}
	where $b_{t,c,s}$ denotes the pixel value at the $s$-th position in the $c$-th channel of $D_t$. We input the channel information into the MLP (${\rm FC}_{(C_{6-t}/2)}-{\rm GELU-Dropout}-{\rm FC}_{(C_{6-t})}$) and then use the Sigmoid to obtain the attention weights $\beta_t$ for each channel of $D_t$,
	\begin{equation}
		\beta_t = {\rm Sigmoid}({\rm MLP}({\rm SoftPool}(D_t))).
	\end{equation}
	To optimize the model training, we also add residual connections and obtain the output
	\begin{equation}
		Y_t={\rm SCA}_t(D_t)=D_t+\beta_t D_t.
	\end{equation}
	The SCA module composed of SoftPool can amplify the channel features related to the prediction and suppress the irrelevant ones while retaining more channel information, thereby further improving the segmentation effect. 

	We represent the output feature maps of each stage of the decoder separately as $Y_1\in {\mathbb R}^{16C\times\frac{H}{16}\times \frac{W}{16}}$, $Y_2\in{\mathbb R}^{8C\times\frac{H}{8}\times \frac{W}{8}}$, $Y_3\in{\mathbb R}^{4C\times\frac{H}{4}\times \frac{W}{4}}$, $Y_4\in{\mathbb R}^{2C\times\frac{H}{2}\times \frac{W}{2}}$, and $Y_5\in{\mathbb R}^{C\times H\times W}$. To complete the segmentation task accurately, we pass $Y_5$ through a $3\times3$ convolution with an output channel of 1 to obtain the pixel-level segmentation result of prediction $Y\in{\mathbb R}^{1\times H\times W}$.	
	
	\subsection{Loss Function}
	Considering the class imbalance issue in medical image segmentation data, we introduce the Tversky loss\cite{tversky} on top of the weighted IoU and weighted BCE losses\cite{wbce}. Therefore, our loss function consists of the weighted IoU loss, weighted BCE loss, and Tversky loss. The overall loss function can be formulated as
	\begin{equation}
		{\mathcal L}={\mathcal L}^W_{IoU}+{\mathcal L}^W_{BCE}+{\mathcal L}_{Tversky},
	\end{equation}
	where the Tversky loss is calculated as follows:
	\begin{equation}
		\resizebox{.85\hsize}{!}{$ \mathcal{L}_{Tversky}=1-\frac{|{\rm GT}\cap {\rm SR}|}{|{\rm GT} \cap {\rm SR}|+\eta|{\rm SR}-{\rm GT}|+\gamma|{\rm GT}-{\rm SR}|}$},
	\end{equation}
	GT and SR denote the ground truth and segmentation result, respectively, $\eta=0.7$, and $\gamma=0.3$.
	
	To avoid difficult training situations, we provide additional supervision to the output of the first and third stages of the decoder\cite{deepsupervised}. Therefore, the final objective function for training can be represented as
	\begin{equation}
		\resizebox{.86\hsize}{!}{${\mathcal L}_{Total}=a{\mathcal L}({\rm GT},Y)+b{\mathcal L}({\rm GT},{\rm head}(Y_1))+c{\mathcal L}({\rm GT},{\rm head}(Y_3)).$}
	\end{equation}
	In the experimental procedure of this study, we set $a=0.6$, $b=0.2$, and $c=0.2$. Here, ${\rm head}(X)$ represents upsampling the input feature map $X$ to directly restore the resolution to $1\times H\times W$.
	
	\section{Experiments and Results}
	We evaluated the performance of HST-MRF in polyp and skin lesion segmentation tasks and compared the results with several related models. To validate the rationality and effectiveness of each component, we conducted ablation experiments in the polyp segmentation task and presented corresponding qualitative results. 
	
	\subsection{Dataset}
	
	\subsubsection{Polyp segmentation}
	In the relevant data set of colonoscopy polyp segmentation, the size and shape of polyps vary greatly. We evaluated our proposed HST-MRF on five publicly available polyp segmentation datasets, including Kvasir-seg\cite{kvasir}, CVC-ClincDB\cite{clinicdb}, ETIS\cite{etis}, EndoScene\cite{endoscene} and ColonDB\cite{colondb}. We followed the experimental settings of DS-TransUNet and conducted independent and cross-study evaluations. Comparative experiments were conducted on two independent datasets, Kvasir-seg and ClinicDB, where the images were resized to $512\times 512$ and $384\times 384$ to ensure comparability of results. Additionally, we performed cross-study evaluation using a training set consisting of 900 and 550 images from Kvasir-seg and CVC-ClinicDB, respectively (a total of 1450 training images). The remaining images from ETIS, EndoScene, CVC-ColonDB, CVC-ClinicDB and Kvasir-seg formed a test set of 798 images.
	
	\subsubsection{Skin lesion segmentation}
	To verify the universality of HST-MRF, we also evaluate its performance on the ISIC 2018 skin lesion segmentation dataset\cite{isic1}, which has 2594 images and their corresponding segmentation labels. Skin lesion segmentation images have low contrast, a wide variety of lesion textures and colors, and may also contain noise (hair and skin texture), among other characteristics. We use fivefold cross-validation in this set of experiments and uniformly resized the images to $256\times256$ to eliminate differences in the original image size.	
	
	\subsection{Implementation and Evaluation Metrics}
	We conducted all experiments using PyTorch on a single NVIDIA RTX 3080 GPU, with the AdamW optimizer (learning rate=1e-4, weight decay=1e-2). The model was trained with a batch size of 16 for 200 cycles, using a dropout rate of 0.5 and an original channel number $C$ of 96. The number of heads in the heterogeneous attention was set to $n_h=4$, and the model's hidden layer dimension was $d=96$. To reduce oscillations in the model, we used cosine annealing to warm up the learning rate.
	
	We used four evaluation metrics: (\rmnum{1}) mean dice coefficient (mDice), (\rmnum{2}) mean intersection over union (mIoU), (\rmnum{3}) recall (Rec.) and (\rmnum{4}) precision (Pre.).
	
	\subsection{Comparison with State-of-the-Art}
	In this section, we present in detail the results of HST-MRF on polyp and skin lesion segmentation in comparison with several excellent models. The models for specific comparisons can be divided into two categories, (\rmnum{1}) CNNs-based models: U-Net\cite{unet}, UNet++\cite{unet++}, R2U-Net\cite{r2u-net}, Double U-Net\cite{DoubleU-Net}, HarDNet-MSEG\cite{HarDNet-MSEG}, FANet\cite{fanet}, Attention U-Net\cite{attentionunet}, PraNet\cite{PraNet}, and Attention R2U-Net\cite{r2u-net}. (\rmnum{2}) Transformer-based models: MCTrans\cite{mctrans}, TransUNet\cite{transunet}, DS-TransUNet\cite{dstransunet}, SETR-PUP\cite{setr}, TransFuse\cite{transfuse}, Swin-Unet\cite{Swin-Unet}, and SegFormer\cite{segformer}.

	\subsubsection{Results of polyp segmentation}
	We conducted comparative experiments on two independent datasets, CVC-ClinicDB and Kvasir-seg, and a cross-study task to validate the effectiveness of HST-MRF in handling polyp segmentation.
	\begin{table*}[h]\tiny
		\centering
		\begin{center}
			\caption{Experimental results comparing HST-MRF with related models on the polyp segmentation Kvasir-seg and CVC-ClinicDB datasets.}
			\label{tab1}
			\resizebox{\textwidth}{!}{
				\begin{tabular}{c| c c c c| c| c c c c}
					\toprule[1pt]
					\multicolumn{5}{c|}{Kvasir-seg} & \multicolumn{5}{c}{CVC-ClinicDB}\\
					\hline
					Method& mDice & mIoU & Rec.& Pre. & Method  & mDice& mIoU& Rec.& Pre.\\
					\hline
					U-Net & 0.783 & 0.684 & 0.808 & 0.828 & U-Net & 0.872 & 0.804 & 0.868 & 0.917\\
					UNet++ & 0.784 & 0.678 & 0.817 & 0.820 & UNet++ & 0.881 & 0.819 & 0.910 & 0.885\\
					Attention U-Net & 0.787 & 0.686 & 0.793 & 0.852 & Attention U-Net & 0.890 & 0.827 & 0.887 & 0.909\\
					Swin-Unet& 0.890 & 0.825 & 0.906 & 0.906  & Swin-Unet & 0.906 & 0.849 & 0.918 & 0.907\\
					DoubleU-Net & 0.813 & 0.733 & 0.840 & 0.861 & DoubleU-Net & 0.924 & 0.861 & 0.846 & \textbf{0.959}\\
					SegFormer& 0.909 & 0.848 & 0.935 & 0.904 & SegFormer & 0.911 & 0.860 & 0.942 & 0.911\\
					HarDNet-MSEG& 0.904 & 0.848 & 0.935 & 0.907 & HarDNet-MSEG & 0.918 & 0.864 & 0.912 & 0.945\\
					MCTrans & 0.862 & - & - & -  & MCTrans & 0.923 & - & - & -\\
					TransUNet& 0.896 & 0.833 & 0.912 & 0.913  & TransUNet & 0.923 & 0.869 & 0.942 & 0.917\\
					FANet& 0.880 & 0.810 & 0.906 & 0.901 & FANet & 0.936 & 0.894 & 0.934 & 0.940\\
					DS-TransUNet-B& 0.911 & 0.856 & 0.935 & 0.914  & DS-TransUNet-B & 0.935 & 0.885 & 0.946 & 0.931\\
					DS-TransUNet-L& 0.913 & 0.859 & 0.936 & \textbf{0.916}  & DS-TransUNet-L & 0.942 & 0.894 & 0.950 & 0.937\\
					\hline
					\textbf{HST-MRF (Ours)}& \textbf{0.914} & \textbf{0.863} & \textbf{0.939} & 0.915 & \textbf{HST-MRF (Ours)}& \textbf{0.949} & \textbf{0.918} & \textbf{0.952} & 0.953\\
					\bottomrule[1pt]
			\end{tabular}}
		\end{center}
	\end{table*}

	\begin{table*}[h]
		\centering
		\begin{center}
			\caption{Comparative experimental results of HST-MRF and correlation models on cross-study tasks based on all polyp segmentation datasets.}
			\label{tab2}
			\resizebox{\textwidth}{!}{
				\begin{tabular}{c|c c c c c c c c c c| c c}
					\toprule[1pt]
					\multirow{2}*{Method} & \multicolumn{2}{|c}{Kvasir-seg} & \multicolumn{2}{c}{ClinicDB} & \multicolumn{2}{c}{ColonDB} & \multicolumn{2}{c}{EndoScene} & \multicolumn{2}{c}{ETIS} & \multicolumn{2}{|c}{Average}\\ 
					~ &mDice&mIoU &mDice&mIoU &mDice&mIoU &mDice&mIoU &mDice&mIoU &mDice&mIoU\\
					\hline
					U-Net &0.818 &0.746 &0.823 &0.755 &0.512 &0.444 &0.398 &0.335 &0.710 &0.626 &0.652 &0.581 \\
					UNet++ &0.821 &0.743 &0.794 &0.729 &0.483 &0.410 &0.401 &0.344 &0.707 &0.624 &0.641 &0.570 \\
					Attention U-Net &0.814 &0.730 &0.850 &0.789 &0.561 &0.484 &0.773 &0.682 &0.371 &0.305 &0.674 &0.598 \\
					PraNet &0.898 &0.840 &0.899 &0.849 &0.709 &0.640 &0.871 &0.797 &0.628 &0.567 &0.800 &0.739 \\
					Swin-Net &0.896 &0.835 &0.899 &0.836 &0.759 &0.666 &0.850 &0.764 &0.681 &0.586 &0.817 &0.737 \\
					SETR-PUP &0.911 &0.854 &0.934 &0.885 &0.773 &0.690 &0.889 &0.814 &0.726 &0.646 &0.847 &0.778 \\
					SegFormer &0.904 &0.844 &0.891 &0.826 &0.762 &0.674 &0.856 &0.780 &0.748 &0.658 &0.832 &0.756 \\
					HarDNet-MSEG &0.912 &0.857 &0.932 &0.882 &0.731 &0.660 &0.887 &0.821 &0.677 &0.613 &0.828 &0.767 \\
					TransUNet &0.912 &0.860 &0.910 &0.856 &0.797 &0.715 &0.887 &0.815 &0.754 &0.671 &0.852 &0.783 \\
					TransFuse-S &0.918 &0.868 &0.918 &0.868 &0.773 &0.696 &0.902 &0.833 &0.733 &0.659 &0.849 &0.786 \\
					TransFuse-L &0.918 &0.868 &0.934 &0.886 &0.744 &0.676 &0.904 &0.838 &0.737 &0.661 &0.847 &0.786 \\
					DS-TransUNet-B&0.934 &0.888 &0.938 &0.891 &0.798 &0.717 &0.882 &0.810 &0.772 &0.698 &0.865 &0.801 \\
					DS-TransUNet-L &\textbf{0.935} &0.889 &0.936 &0.887 &0.798 &0.722 &\textbf{0.911} &\textbf{0.846} &0.761 &0.687 &0.868 &0.806 \\
					\hline
					\textbf{HST-MRF (Ours)} & \textbf{0.935} & \textbf{0.894} & \textbf{0.940} & \textbf{0.905} & \textbf{0.833}& \textbf{0.774} & 0.902 & \textbf{0.846} & \textbf{0.778} & \textbf{0.725} & \textbf{0.879} & \textbf{0.829}\\
					\bottomrule[1pt]
			\end{tabular}}
		\end{center}
	\end{table*}

	We evaluated HST-MRF on Kvasir-seg and CVC-ClinicDB by comparing mDice, mIoU, recall, and precision of each model. The detailed results are shown in \textcolor{mblue}{Table \ref{tab1}}. \textcolor{mblue}{Table \ref{tab2}} shows the test results of mDice and mIoU of HST-MRF and related models on the cross-study task of polyp segmentation. We also added an average term to evaluate the model's overall performance. In accordance with \textcolor{mblue}{Tables \ref{tab1}} and \textcolor{mblue}{{\ref{tab2}}}, the following observations can be made.
	
	Compared with U-Net, its variants can achieve even better results. For example, on Kvasir-seg, FANet's mDice and mIoU reach 0.88 (mDice=0.783 for U-Net) and 0.901 (mIoU=0.810 for U-Net), respectively, and HarDNet-MSEG (mDice and mIoU on average are 0.828 and 0.767) outperforms U-Net (mDice of 0.652 and mIoU of 0.581 on Average) on the cross-study task. This finding indicates that the U-Net structure still has more room for expansion and optimization.
	
	Transformer-based models, such as DS-TransUNet and TransFuse, performed remarkably better than U-Net and its CNNs variants on polyp segmentation. This finding suggests that Transformer can compensate for the inherent shortcomings of traditional CNNs and achieve higher accuracy in medical image segmentation. Among them, DS-TransUNet performed the best in independent and cross-study evaluations (excluding HST-MRF). This condition may be attributed to the dual-branch structure of DS-TransUNet, which reduces information loss caused by patch segmentation.
	
	In accordance with \textcolor{mblue}{Table \ref{tab1}}, HST-MRF performs worse in precision on two sets of data, one is 0.1 percentage points lower and one is 0.6 percentage points worse. However, HST-MRF can achieve better results in the remaining evaluation metrics, especially the mIoU of CVC-ClinicDB, which exceeds the second place by 2.4 percentage points, and the improvement is huge.
	
	In accordance with \textcolor{mblue}{Table \ref{tab2}}, the effect of HST-MRF on EndoScene segmentation is not outstanding, with a corresponding mIoU of 0.902, which is lower than DS-TransUNet-L and TransFuse-L. However, HST-MRF demonstrates excellent performance on the ETIS and ColonDB datasets, with mDice improving by about 4 percentage points and mIoU improving by 5 percentage points in ColonDB, and mIoU improving by about 4 percentage points in ETIS. Overall, the evaluation metrics in the Average category show improvements of at least 1 percentage point.
	
	\subsubsection{Results of skin lesion segmentation}
	The evaluation results of HST-MRF on the skin lesion segmentation ISIC 2018 dataset are shown in \textcolor{mblue}{Table \ref{tab3}}, with mDice, mIoU, recall, and precision used as evaluation metrics. 
	
	In accordance with \textcolor{mblue}{Table \ref{tab3}}, CNN-based models perform worse than Transformer-based models in skin lesion segmentation. The dual-branch structure of DS-TransUNet-L achieved the highest recall of 0.922 and outperformed other models in other metrics (except for HST-MRF). This finding further demonstrates the advantages of the dual-branch structure of DS-TransUNet.
	
	In addition to recall, HST-MRF achieved the best performance on other metrics, with its mIoU and precision exceeding the second-best by around 2 percentage points, reaching 0.870 and 0.947, respectively. 	
	
	HST-MRF performs well in polyp segmentation and achieves good results in skin lesion segmentation by analyzing the above experimental results, demonstrating its good generalization ability. In combination with the performance of DS-TransUNet, this indicates that addressing the issue of structural information loss caused by patch segmentation is necessary.	
		
	\begin{table}[h]
		\centering
		\begin{center}
			\caption{Comparative experimental results of HST-MRF and correlation models on skin lesion segmentation data ISIC 2018.}
			\label{tab3}	
			\resizebox{0.9\columnwidth}{!}{
				\begin{tabular}{c|c c c c}
					\toprule[1pt]
					Method & mDice & mIoU & Rec. & Pre.\\
					\hline
					U-Net & 0.674 & 0.549 & 0.708 &-\\
					AttentionU-Net & 0.665 & 0.566 & 0.717 &-\\
					R2U-Net & 0.679 & 0.581 & 0.792 & -\\
					Attention R2U-Net & 0.691 &0.592&0.726&-\\
					FANet& 0.873&0.802&0.865&0.924\\
					DoubleU-Net& 0.896& 0.821&0.878&0.946\\
					Swin-Unet& 0.897&0.829&0.903&0.920\\
					SegFormet& 0.902&0.836&0.911&0.921\\
					TransUNet& 0.906&0.841&0.913&0.923\\
					DS-TransUNet-B&0.910 &0.848&0.911&0.934\\
					DS-TransUNet-L&0.913 &0.852&\textbf{0.922}&0.927\\
					\hline
					\textbf{HST-MRF (Ours)} &\textbf{0.919} &\textbf{0.870} &0.913 &\textbf{0.947}\\
					\bottomrule[1pt]
			\end{tabular}}
		\end{center}
	\end{table}
	
	\subsection{Ablation Study}
	We conducted ablation experiments on Kvasir-seg and CVC-ClinicDB datasets to verify (\rmnum{1}) whether each module can function as intended and (\rmnum{2}) whether multi-receptive field patch segmentation can alleviate the structural information loss problem. To this end, we defined seven states for HST-MRF: (1) No APE: discarding the APE module and using normal patch flattening embedding. (2) No HST: heterogeneous attention is not used, and instead, independent parallel branching structures are used to extract information from different receptive fields. (3) No MBP: removing the MBP module and only using normal skip connection. Specifically, the encoded and decoded feature map information of two different receptive fields are directly connected. (4) No SCA: removing the SCA module from HST-MRF, that is, no channel attention is used. (5) Max+Avg CA: using max + average pooling to build channel attention. (6) One Receptive Field (abbreviated as One RF): dividing the image into patches only under a fixed receptive field, and using Swin Transformer to model the long-range relationship and MBP to fuse the encoding-decoding information. (7) HST-MRF: representing the complete state, as shown in \textcolor{mblue}{Fig. \ref{fig1}}.
	
	\begin{table}[h]
		\centering
		\begin{center}
			\caption{Ablation experiment results of HST-MRF on polyp segmentation Kvasir-seg and CVC-ClinicDB datasets.}
			\label{tab4}
			\resizebox{0.9\columnwidth}{!}{	
				\begin{tabular}{c|c c|c c}
					\toprule[1pt]
					\multirow{2}*{Method} &\multicolumn{2}{c|}{Kvasir} & \multicolumn{2}{|c}{CVC-ClinicDB}\\
					~& mDice & mIoU  & mDice & mIoU \\
					\hline
					No APE &0.908 &0.856 &0.941 &0.907\\
					No HST &0.891 &0.833 & 0.931 &0.890\\
					No MBP & 0.893 &0.836 &0.910 &0.866\\
					No SCA & 0.899 &0.844 &0.923 &0.880\\
					Max+Avg CA & 0.902 & 0.847 &0.938 &0.901\\
					One RF &0.877 &0.816&0.897 &0.847\\
					HST-MRF &\textbf{0.914} &\textbf{0.863} &\textbf{0.949} &\textbf{0.918}\\
					\toprule[1pt]
			\end{tabular}}	
		\end{center}
	\end{table}

	The results are presented in \textcolor{mblue}{Table \ref{tab4}}. We randomly selected challenging segmentation instances from Kvasir-seg and CVC-ClinicDB for visualization to better observe the results of the ablation study. The specific qualitative results are shown in \textcolor{mblue}{Fig. \ref{fig6}}, where green indicates ground truth and red indicates predicted segmentation. It is worth noting that the instances in \textcolor{mblue}{Fig. \ref{fig6}} were randomly selected, and although they may not be representative of the model's performance on the entire dataset, they can still provide valuable insights.
	
	\subsubsection{Verification of module validity}
	On the basis of the test results in \textcolor{mblue}{Table \ref{tab4}} and the visualization in \textcolor{mblue}{Fig. \ref{fig6}}, the following observations can be made.
	
	Although APE has the least effect on HST-MRF, it still leads to performance improvement. This finding indicates that compared with the ordinary flattened embedding, the APE module is designed to handle different regions of the image more flexibly, and the patch embedding contains more valuable information.
	
	No HST uses two parallel branches to process information from different receptive fields and merges the information using the MBP module. The results indicate that HST-MFT outperforms No HST, indicating that using HST can effectively guide multi-receptive field interactive learning at each stage of encoding, and transfer it to the next stage for progressive learning, thereby achieving the complementary integration of structural information.

	Removing the MBP module remarkably decreases the performance of HST-MRF and MBP can more effectively fuse low-level and high-level features compared with skip connection. Combining MBP and HST promotes the complementary of multi-receptive field information and minimizes the loss of structural information.
	
	The performance of HST-MRF, Max+Avg CA, and No SCA decreases on the two sets of data in order. We can draw the following conclusions. The channel attention module design can indeed solve the problem of channel redundancy. The SCA, compared to the ordinary channel attention module, can retain more relevant channel information, thereby improving the segmentation quality.
	
	In summary, the absence of any of the modules will cause a degradation in HST-MRF performance, although a clear difference is observed in the contribution of different components to performance.
	
	\subsubsection{Validation of the effectiveness of multi-receptive field patch segmentation}
	In accordance with \textcolor{mblue}{Table \ref{tab4}}, the one receptive field (One RF) state performed poorly in the polyp segmentation. Comparing the results of HST-MRF, No HST, No MBP, and One RF, models considering multiple receptive fields outperformed those considering only one. This finding indicates that multi-receptive field patch segmentation has certain advantages. Moreover, the performance of the models decreased in the order of HST-MRF, No HST, No MBP, and One RF. This indicates that the combination of HST and MBP to fuse information from multi-receptive fields is the most effective, followed by HST alone and MBP alone.
	
\begin{figure}
		\includegraphics[width=\columnwidth]{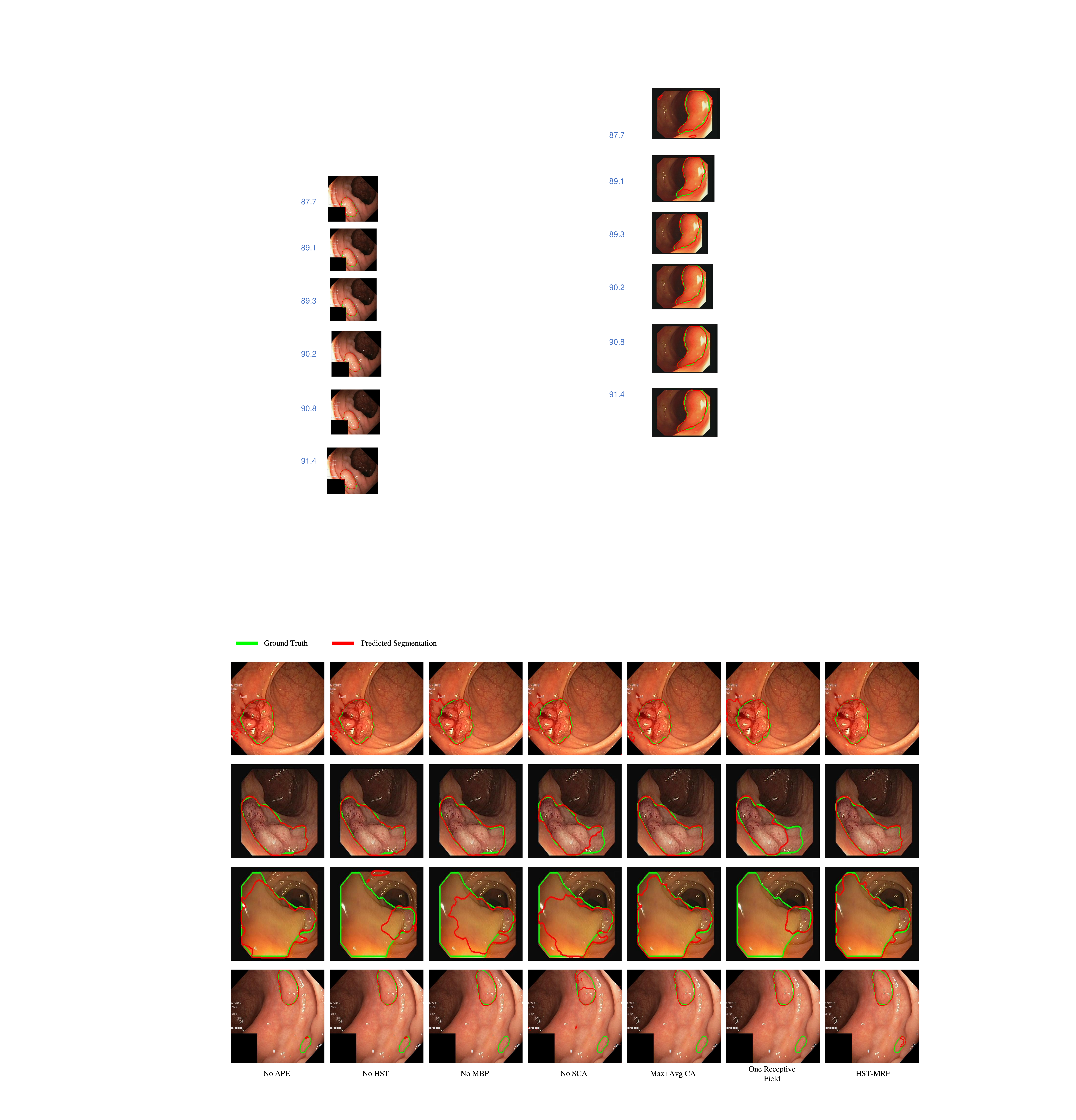}
		\caption{Qualitative results for randomly selected instances in the ablation study. The \textcolor{green}{green} and \textcolor{red}{red} segmentation lines in the visualization represent ground truth and predicted segmentation, respectively.}
		\label{fig6}
	\end{figure}

	Similar observations can be made from \textcolor{mblue}{Fig. \ref{fig6}}. In the first three instances of \textcolor{mblue}{Fig. \ref{fig6}}, the area to be segmented is highly irregular, and the model's segmentation performance is poorest in one RF, whereas HST-MRF achieved a segmentation boundary that closely matched the ground truth. In the fourth instance (bottom row), two areas need to be segmented, but only HST-MRF detected a portion of them. These results indicate the advantages of multi-receptive field patch segmentation. The use of multiple receptive fields can effectively reduce the loss of structural information, potentially improving the performance of the model in medical image segmentation tasks. Additionally, the combination of our designed HST and MBP with multi-receptive field same-scale patch segmentation minimizes the loss of structural information and enables more accurate pixel-level segmentation.

	\section{Discussion and Conclusion}
	Transformer is a powerful model for modeling long-range dependencies between visual elements. However, their reliance on patch segmentation can result in a loss of structural information, which is especially problematic in medical image segmentation. Therefore, reducing the loss of structural information when using vision-based Transformer models is crucial. To address this problem, we propose HST-MRF, which outperforms comparative methods on polyp and skin lesion segmentation tasks.
	
	In the past, limited has been paid to the issue of structural information loss when using Transformer in computer vision. Compared with these approaches, HST-MRF considers the use of heterogeneous attention to effectively combine patch information from different receptive fields and incorporates the HST module to address this problem. This module's concept is applicable to image segmentation and can be extended to other tasks like image classification and detection. However, using HST requires us to extract different receptive field information from the image, which may result in a loss of some information. Directly performing segmentation at different scales on the original image can compensate for this problem and one of our future research directions is to explore how to apply HST to the combination of multiscale patches.
	
	In this work, we also designed the MBP module to assist HST in further fusing different receptive field information and reducing semantic confusion when fusing encoding and decoding information. HST and MBP have good scalability and can be extended to multi-receptive fields (more than two types) for more complex segmentation tasks. In addition, APE and SCA further improve segmentation quality through adaptive calculation. Although the improvement effect is not significant,exploring how to flexibly handle different regions of the image to improve performance is valuable.
	
	In this study, we propose a HST-MRF model based on a U-Net encoder-decoder architecture for medical image segmentation, aiming to address the problem of structural information loss caused by patch-based segmentation. We introduced heterogeneous attention and designed an interactive encoding mechanism. The HST module achieves interaction between different receptive fields information and transmit it to the next stage, completing progressive learning, thereby reducing structural information loss. We proposed a two-stage fusion MBP module to assist HST in further fusing encoding information, connecting low-level and high-level feature information to accurately locate the lesion area. To achieve higher segmentation quality, we also designed APE and SCA modules to preserve more useful information based on adaptive computation in ordinary patch embedding and channel feature selection. HST-MRF outperforms previous state-of-the-art methods on relevant datasets for polyp and skin lesion segmentation tasks. The effectiveness of the modules and the multi-receptive field segmentation of patches of the same scale were verified through ablation experiments. In the future, we will extend HST-MRF to handle more complex medical image segmentation tasks.

	{\small
		\bibliographystyle{IEEEtran}
		\bibliography{IEEEabrv, references}
	}
	
\end{document}